\documentclass{article}

\PassOptionsToPackage{numbers,compress}{natbib}
\usepackage{microtype}
\usepackage{graphicx}
\usepackage{xcolor}
\usepackage{subcaption}
\usepackage{amsmath}
\usepackage{algorithm}
\usepackage{algorithmic}
\usepackage{xcolor}
\usepackage{booktabs} 
\usepackage{hyperref}
\usepackage{threeparttable} 
\usepackage{array}          
\usepackage{makecell}       
\usepackage{xcolor}         
\usepackage{amsfonts}
\usepackage{graphicx}        
\usepackage{multirow}        
\usepackage[table]{xcolor}
\usepackage{subcaption}
\usepackage[preprint]{neurips_2026}
\usepackage[utf8]{inputenc} 
\usepackage[T1]{fontenc}    
\usepackage{hyperref}       
\usepackage{url}            
\usepackage{booktabs}       
\usepackage{amsfonts}       
\usepackage{nicefrac}       
\usepackage{microtype}      
\usepackage{xcolor}         

\title{KVCapsule: Efficient Sequential KV Cache Compression for Vision-Language Models with Asymmetric Redundancy}

\author{
  Yingbing Huang$^{1}$ \quad  Tharun Adithya Srikrishnan$^{2}$ \quad Steven K. Reinhardt$^{2}$ \quad Deming Chen$^{1}$ \\
  $^{1}$ University of Illinois Urbana-Champaign \quad $^{2}$ AMD\\
}

\begin{document}

\maketitle

\def\kv{\textcolor{black}{KVCapsule}}
\begin{abstract}
  Vision–Language Models (VLMs) have emerged as a critical and fast-growing extension of Large Language Models (LLMs) that enable multimodal reasoning through both text and image inputs. Although VLMs enrich the capabilities of language models, they also inherit and amplify key computational bottlenecks: the memory overhead caused by the large key–value (KV) cache during autoregressive decoding. This challenge is particularly severe in VLMs, where images produce longer token sequences and denser feature representations compared to text. Moreover, the spatial and information-rich nature of vision tokens introduces structured attention patterns that make many LLM-oriented KV cache compression techniques ineffective when applied directly to VLMs.
  
  In this work, we conduct a detailed empirical analysis of the behavior of vision tokens, highlighting the critical differences from purely text-based models. Based on these insights, we propose \kv, a novel KV cache compression framework for vision tokens. \kv\ keeps the pretrained VLM backbone frozen, requires no modification to the attention computation modules, and can be integrated into existing VLMs through lightweight compression and reconstruction components. We evaluate \kv\ on multiple VLMs and benchmark tasks, demonstrating up to $2 \times$ improvement in TPS and $2.4\times$ reduction in KV cache memory at a 60\% compression ratio, with negligible degradation in accuracy or response quality. Our findings offer practical pathways to scale VLM inference under constrained memory budgets and inspire further research into structure-aware cache compression for multimodal models.
\end{abstract}

\begin{figure*}[ht]
\begin{center}
\centerline{\includegraphics[width=\linewidth]{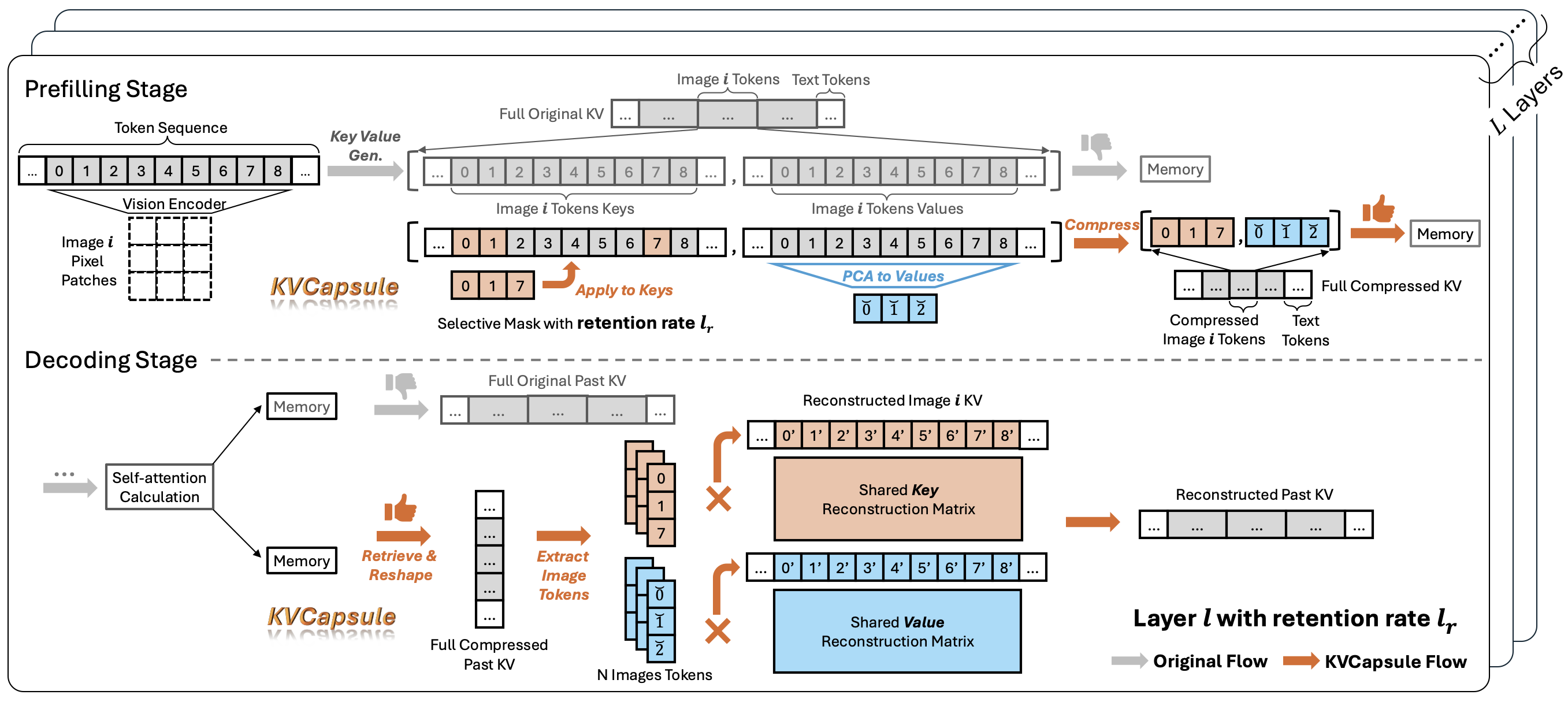}}
\vspace{-5pt}
\caption{\textbf{Overview of the \kv\ framework.} The gray region denotes the original process of VLMs, while the yellow region shows the process of \kv.}
\vspace{-30pt}
\label{fig:flow}
\end{center}
\end{figure*}

\section{Introduction}
\label{sec:introduction}
Extending LLMs to the multimodal domain has led to rapid progress in VLMs, enabling applications such as image captioning, visual question answering, and multimodal dialogue. Recent open-source state-of-the-art VLMs, including Qwen-VL~\cite{Qwen-VL}, LLaVA-NeXT~\cite{liu2024llavanext}, and InternVL~\cite{zhu2025internvl3}, demonstrate increasingly strong multimodal reasoning capabilities.

Despite this progress, VLMs face a major scalability bottleneck during autoregressive decoding: the memory required to store and move the KV cache. This issue is particularly severe for vision tokens, which are encoded at higher granularity than text~\cite{li2025madakv, huang2025aircache}. A single image is typically represented by hundreds or thousands of tokens, so even a few images can substantially increase the effective context length and rapidly exhaust GPU memory.

This challenge is amplified in multi-image and long-context settings that arise naturally in practice, such as visual comparison, sequential reasoning over image sequences, video understanding via frame sampling, and multi-turn multimodal dialogue~\cite{alayrac2022flamingo, zhang2023video, suhr2019corpus}. In these scenarios, vision tokens dominate the input sequence and persist in the KV cache throughout decoding, causing memory consumption to scale with both the visual context length and the decoding length. Efficient KV cache reduction is therefore essential for scalable VLM inference under realistic hardware constraints.

To mitigate KV cache growth, prior work has explored compression techniques including quantization~\cite{hooper2024kvquant, cheng2025qaq}, low-rank approximation~\cite{zhang2024lorc, chang2025palu}, and attention-based pruning~\cite{li2024snapkv, wan2024look}, with recent extensions to VLMs~\cite{su2025akvq, han2025calibquant, saxena2024eigen, mu2025sals, tu2024vl, chang2025xkv, pei2024cross, jiang2025purekv, yang2025topv, huang2025aircache}. These approaches typically compress the visual KV cache using attention statistics from prefilling, assuming that token importance remains stable during generation.

However, this assumption is inherited from text-only LLMs and does not necessarily hold for vision tokens. In VLMs, visual attention can vary substantially across layers, heads, and decoding steps, especially in long-context and multi-image settings, making static or attention-predicted compression policies unreliable.

Our analysis reveals that vision token importance is highly context-dependent during decoding. Different layers and heads attend to different visual regions, and the set of important tokens can change as generation progresses from local visual grounding to more semantic reasoning. This behavior makes static compression fragile: a token that receives little attention during prefilling may still become useful in later decoding steps, especially in long-context or multi-image inputs where the model must revisit different parts of the visual context.

Motivated by this observation, we propose \kv, a backbone-frozen KV cache compression framework for vision tokens. Rather than permanently removing low-ranked tokens, \kv\ stores compact sequence-level representations of the visual KV cache before attention score computation, as shown in Figure~\ref{fig:flow}. Importantly, \kv\ uses asymmetric compression for keys and values: it reconstructs keys to preserve attention geometry, while compressing values along the sequence dimension to retain their dominant semantic subspace. At decoding time, these compressed states are reconstructed when needed, so attention can still be computed over the evolving visual context. This design separates memory reduction from irreversible token dropping, preserving flexibility under changing attention patterns while adding only lightweight matrix-multiplication overhead.

\kv\ integrates seamlessly into existing VLMs without requiring architecture changes or retraining. Evaluated across multiple VLM architectures and benchmark datasets, our method achieves substantial memory savings with minimal accuracy loss. We summarize the key contributions of this work as follows:
\begin{itemize}

    \item \textbf{Sequential redundancy discovery.} We identify a high degree of sequential redundancy in visual KV cache; this redundancy accumulates with increasing sequence length and is overlooked by current compression techniques (Section \ref{sec:temp_rendundancy}).

    \item \textbf{Attention dynamics analysis.} We show that vision-token attention varies across layers, heads, and decoding steps, challenging static importance assumptions (Section~\ref{sec:temp_dynamics}).

    \item \textbf{Asymmetric KV redundancy.} We empirically demonstrate a structural disparity between visual keys and values. This observation motivates our asymmetric strategy for KV cache compression (Section \ref{sec:asym}).

    \item \textbf{Sequential KV compression.} We introduce \kv, which compresses the vision-token KV cache across sequence and reconstructs it before attention computation to support dynamic attention (Section \ref{sec:compression_reconstruction}).

    \item \textbf{Simple integration.} \kv\ leaves the VLM backbone and attention modules unchanged, adding only lightweight compression and reconstruction components (Section~\ref{sec:inference}).

    \item \textbf{Efficient inference.} We show that \kv\ substantially reduces KV memory while preserving accuracy, enabling scalable long-context and multi-image reasoning (Section~\ref{sec:vlmevalkit} \& \ref{sec:video_evaluation}). We further introduce fused \kv, which integrates reconstruction into attention to reduce runtime overhead and improve decoding throughput (Section~\ref{sec:efficiency_analysis}).
\end{itemize}

\section{Observations}
In this section, we analyze the structure and decoding behavior of visual KV states in VLMs using 500 randomly sampled examples from MEGABench~\cite{chen2025mega-bench}. Our measurements reveal three design-driving properties: sequence-level redundancy among vision tokens, dynamic vision token importance during decoding, and distinct structural characteristics of keys and values. These findings expose a mismatch between visual KV behavior and static-importance pruning methods designed for text-only LLMs.

\subsection{Sequential Redundancy}
\label{sec:temp_rendundancy}
\begin{figure}[t!]
\centering

\begin{minipage}[t]{0.50\columnwidth}
    \vspace{0pt}
    \centering
    \captionof{figure}{\textbf{Prefilling attention structure.}
    Attention forms image-level blocks and selective cross-modal patterns over vision tokens.}
    \label{fig:pattern_kv}
    \vspace{-4pt}
    \includegraphics[width=0.92\linewidth]{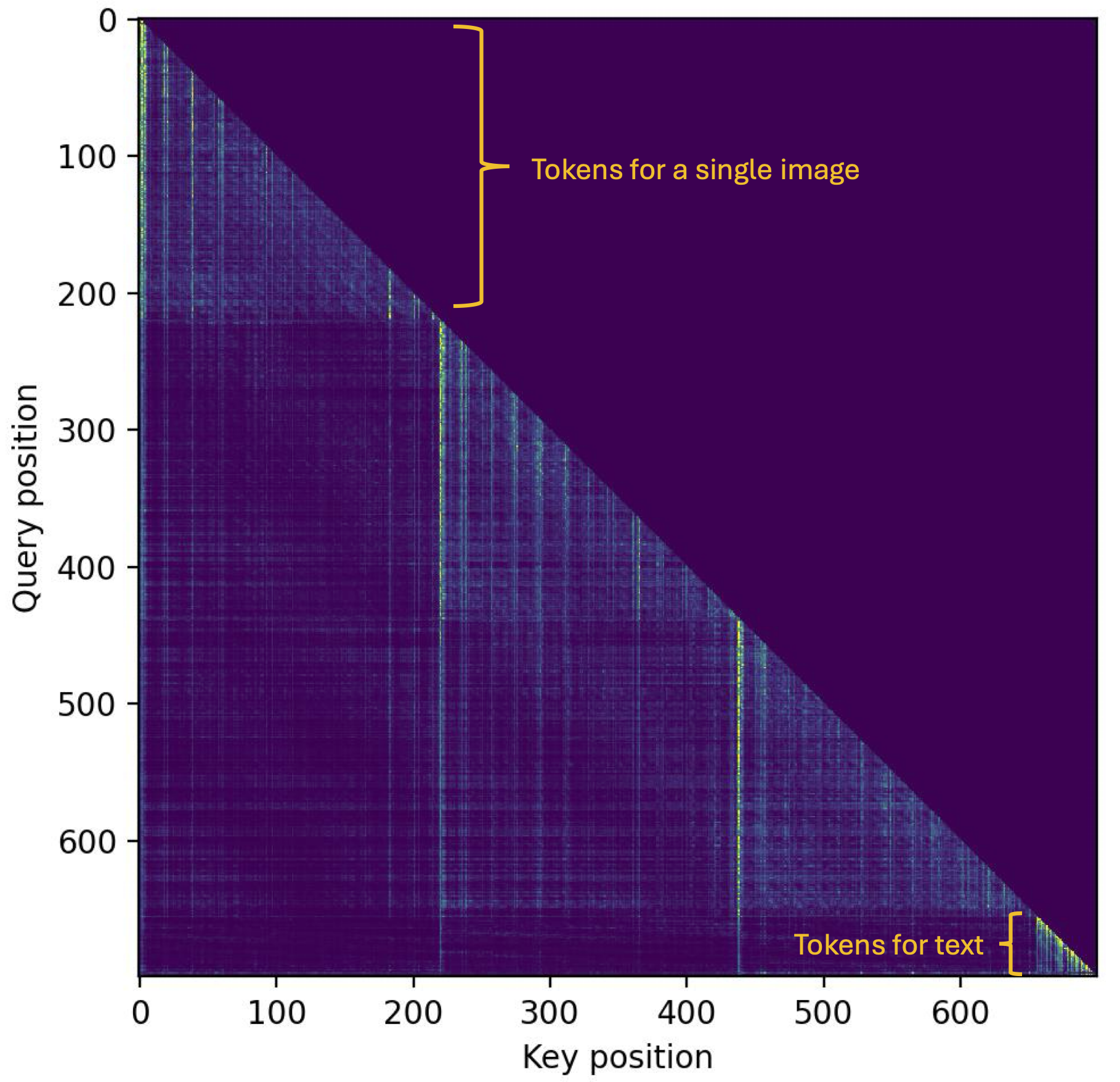}
\end{minipage}
\hfill
\begin{minipage}[t]{0.47\columnwidth}
    \vspace{0pt}
    \centering
    \captionof{table}{Top-10\% attended vision-token overlap across decoding steps. Each entry reports the overlap between the top-attended vision-token sets at reference step $i$ and comparison step $j$.}
    \label{tab:top_visual_token_overlap}
    \vspace{-4pt}
    \small
    \setlength{\tabcolsep}{4.2pt}
    \renewcommand{\arraystretch}{1.22}
    \begin{tabular}{c|ccccc}
    \specialrule{1pt}{0pt}{2pt}
    \makecell{\textbf{Step}\\\textbf{$j$}}
    & \multicolumn{5}{c}{\textbf{Reference step $i$}} \\
    \cline{2-6}
    & \textbf{0}
    & \textbf{1}
    & \textbf{5}
    & \textbf{10}
    & \textbf{20} \\
    \specialrule{1pt}{2pt}{2pt}

    1  & 0.42 & --   & --   & --   & --   \\
    5  & 0.16 & 0.33 & --   & --   & --   \\
    10 & 0.16 & 0.42 & 0.42 & --   & --   \\
    20 & 0.25 & 0.42 & 0.50 & 0.54 & --   \\
    40 & 0.25 & 0.33 & 0.42 & 0.71 & 0.66 \\
    60 & 0.25 & 0.33 & 0.42 & 0.58 & 0.46 \\

    \midrule
    \makecell{\textbf{Avg.}\\\textbf{overlap}}
       & 0.25 & 0.37 & 0.44 & 0.61 & 0.56 \\

    \specialrule{1pt}{2pt}{0pt}
    \end{tabular}
\end{minipage}

\vspace{-8pt}
\end{figure}

Visual inputs contain spatial correlations and object-level coherence, creating redundancy across vision tokens. This redundancy is already visible during prefilling, when the multimodal input is encoded into the KV cache. As shown in Figure~\ref{fig:pattern_kv}, attention forms image-level blocks induced by spatial coherence, while the instruction selectively attends to a compact set of salient visual regions. This structured attention indicates redundant token-level information in visual KV states, making sequence-level compression a natural way to reduce KV memory.

\subsection{Sequential Dynamics of Vision-Token Importance}
\label{sec:temp_dynamics}

Structured redundancy does not imply static importance. During autoregressive decoding, the model may require different visual evidence for different generated tokens, making prefilling-only or early-decoding importance estimates unreliable.

\begin{figure}[t!]
\centering

\begin{minipage}[t]{0.49\columnwidth}
    \vspace{0pt}
    \centering
    \includegraphics[height=0.34\textheight,keepaspectratio]{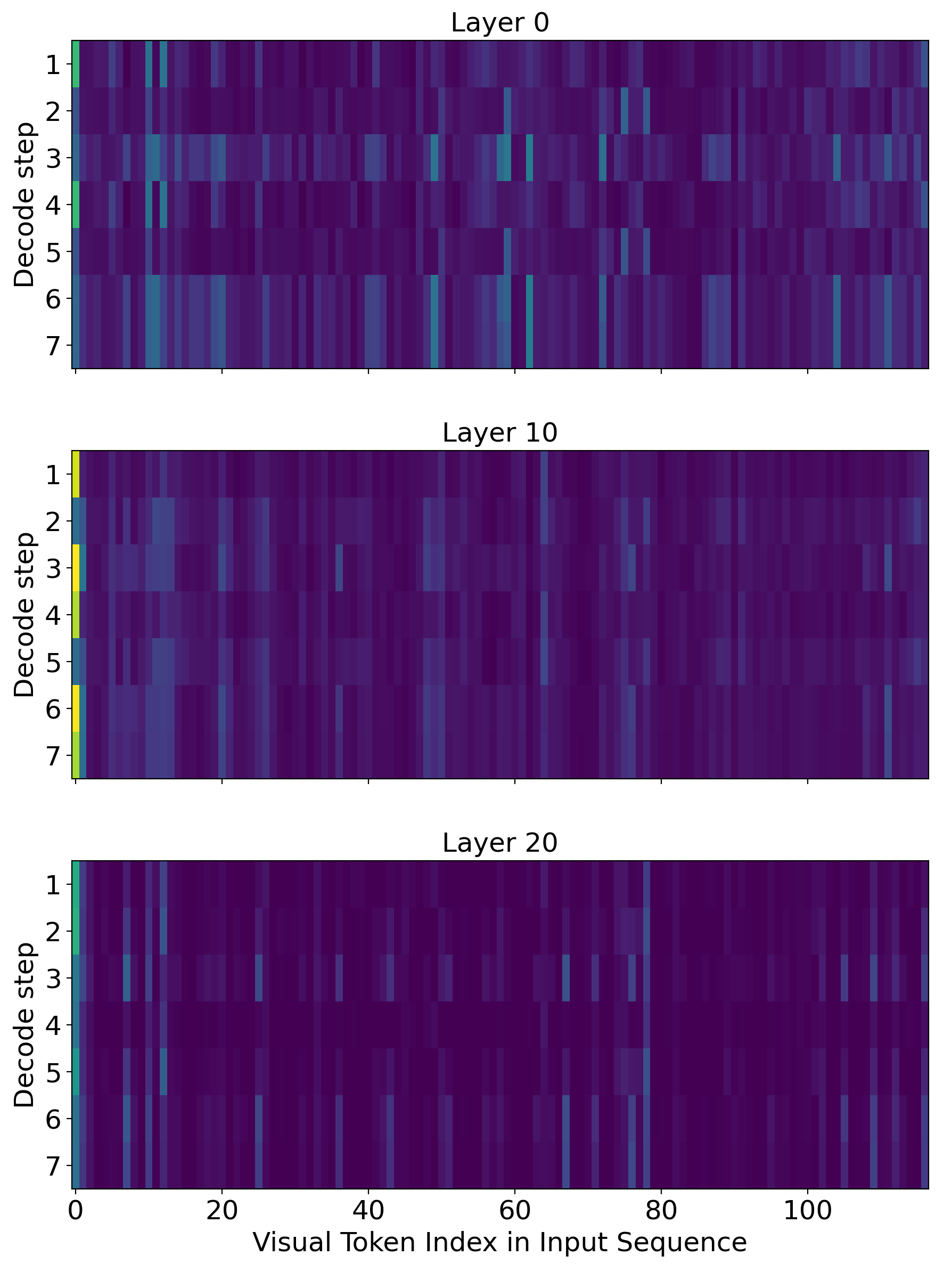}
    \vspace{-4pt}
    \captionof{figure}{\textbf{Vision-token attention dynamics.} Attention shifts across decoding steps and layers; brighter colors indicate higher attention over vision tokens.}
    \label{fig:dynamic_kv}
\end{minipage}
\hfill
\begin{minipage}[t]{0.49\columnwidth}
    \vspace{0pt}
    \centering
    \includegraphics[height=0.34\textheight,keepaspectratio]{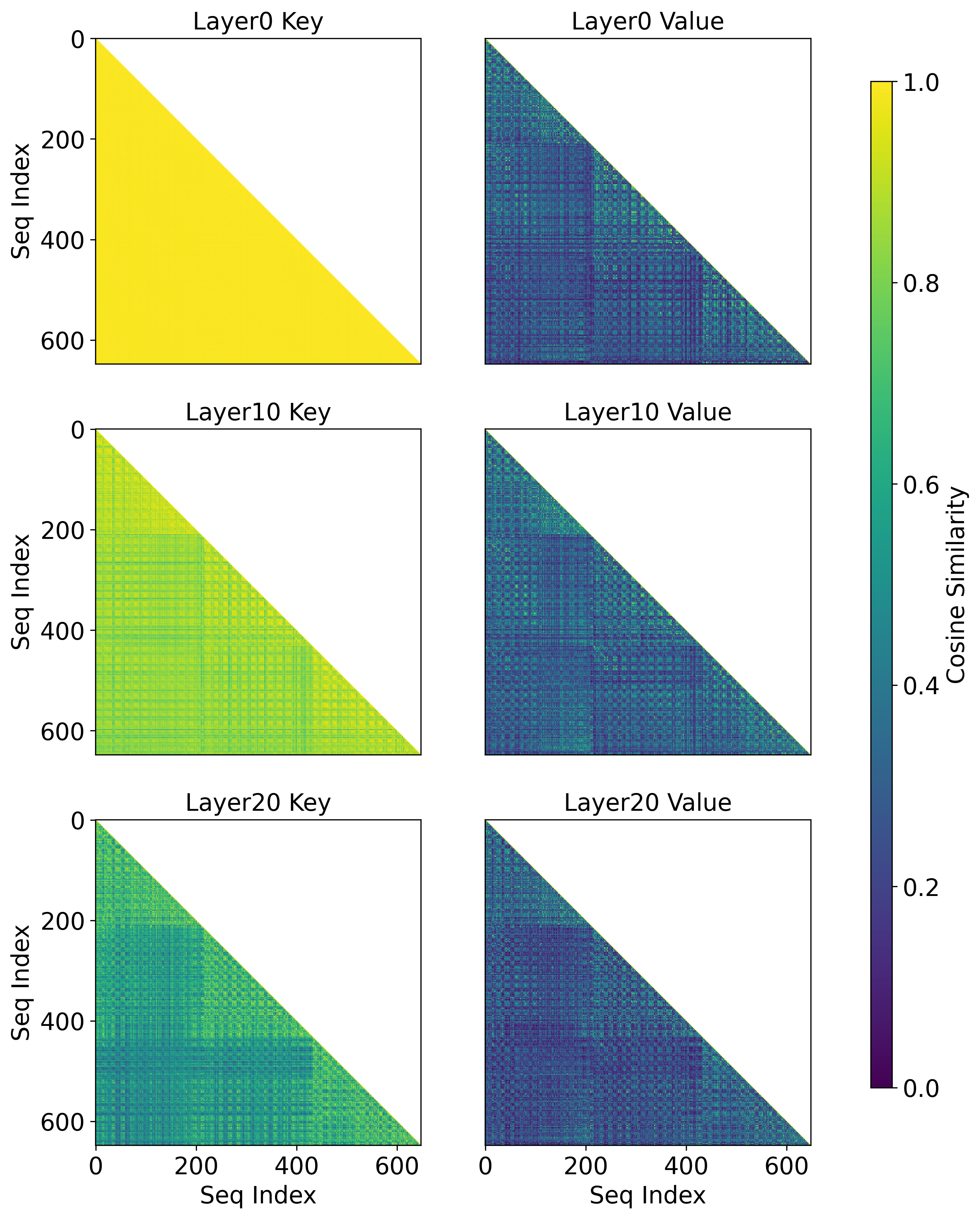}
    \vspace{-4pt}
    \captionof{figure}{\textbf{Key--value similarity across layers.} Keys show stronger early-layer redundancy, while values are more token-specific, motivating asymmetric compression.}
    \label{fig:kv_cos_sim}
\end{minipage}

\vspace{-10pt}
\end{figure}

Figure~\ref{fig:dynamic_kv} shows that attention over vision tokens shifts across decoding steps and layers: salient tokens may fade, while previously suppressed tokens can re-emerge. Table~\ref{tab:top_visual_token_overlap} quantifies this shift using the overlap of top-10\% attended vision-token sets. The overlap between step 0 and steps 5 or 10 is only 0.16, and the average overlap with step 0 is only 0.25; even later-step overlaps remain far from complete.

This dynamic behavior makes static token pruning risky, since removed tokens cannot contribute if they become important later. To avoid this irreversible loss, \kv\ introduces a reconstructable visual KV cache: visual KV states are stored in compressed form and reconstructed to full length before attention, allowing the model to adapt as vision token importance shifts during generation.

\subsection{Asymmetric Redundancy in KV States}
\label{sec:asym}

Beyond decoding dynamics, visual keys and values exhibit distinct sequence-level structures, suggesting that they should not be compressed with the same mechanism. To characterize this asymmetry, we measure pairwise cosine similarity between key and value vectors along the vision-token sequence dimension. As shown in Figure~\ref{fig:kv_cos_sim}, keys have substantially higher inter-token similarity than values, especially in early layers.

This key redundancy indicates that many visual keys encode similar query-matching geometry. Since keys affect attention through the query-key product, a compact subset of representative keys can preserve much of the attention structure. However, because vision token importance changes during decoding, permanently pruning the remaining keys can still remove information needed later. We therefore retain representative keys and reconstruct the full key sequence before attention, reducing persistent KV storage while preserving recoverable query-key geometry.

Values exhibit much lower pairwise similarity, indicating that individual value tokens carry more token-specific visual content. Directly pruning value rows can therefore discard localized semantic information. However, low pairwise similarity does not rule out low-rank structure of the full value matrix: pairwise cosine similarity measures local token-to-token closeness, whereas PCA captures dominant global modes of variation across the sequence. This motivates sequence-level PCA for values, which compresses along the token dimension while preserving contributions from all vision tokens. Appendix~\ref{app:hidden_dim_pca} further shows that visual values are dense along the hidden dimension, supporting sequence-level compression over hidden-dimension latent compression.

This observation motivates the asymmetric design of \kv: keys are selectively retained and reconstructed to preserve attention geometry, while values are compressed by sequence-level PCA to preserve distributed semantic content without dropping entire tokens. Appendix~\ref{app:kv_redundancy_analysis} provides quantitative evidence for this key--value distinction, and Appendix~\ref{app:key_value_reconstruction} supports the MLP-for-keys/PCA-for-values design.

\subsection{Layer-wise Compression Tolerance}
\label{sec:layer_wise}

Figure~\ref{fig:kv_cos_sim} shows that visual KV redundancy is layer-dependent: early-layer keys exhibit stronger global similarity, while deeper-layer keys become more structured and less uniformly redundant; values remain comparatively token-specific across layers. Since compression errors in shallow layers can propagate through many subsequent layers, a uniform compression ratio is suboptimal. \kv\ therefore adopts a pyramid retention schedule, preserving more visual KV information in earlier layers while applying stronger compression in later layers. Appendix~\ref{app:compression_ratio} validates this design through a layer-wise compression sweep; for example, Layer~35 maintains cosine similarity above 0.93 even at a 95\% compression rate.

\section{\kv\ Framework}
\label{sec:framework}

\subsection{Learning Compression and Reconstruction}

\label{sec:compression_reconstruction}

The observations above motivate \kv’s asymmetric, recoverable design: keys are selectively retained and reconstructed to preserve attention geometry, while values are compressed by sequence-level PCA to preserve distributed semantic content without irreversible token pruning.

In standard multi-head attention, the attention output for queries $\mathbf{Q} \in \mathbb{R}^{s \times d_\text{head}}$ is computed as
\begin{equation*}
\text{Attention}(\mathbf{Q}, \mathbf{K}, \mathbf{V}) =
\mathrm{softmax}\!\left(\frac{\mathbf{Q}\mathbf{K}^\top}{\sqrt{d_\text{head}}}\right)\mathbf{V},
\label{eq:1}
\end{equation*}
where $\mathbf{K} \in \mathbb{R}^{m \times d_\text{head}}$ and $\mathbf{V} \in \mathbb{R}^{m \times d_\text{head}}$ denote the cached keys and values, and $m$ is the sequence length.
The query-key product determines attention weights, while the attention-value product aggregates semantic content into the output. This functional distinction motivates different compression strategies for keys and values.

\paragraph{Selective Retention and Reconstruction for Keys.}

Keys determine attention weights through their similarity to queries. Therefore, key compression should preserve the geometry of the query-key product rather than merely retain tokens with high static attention scores. Motivated by the high inter-token redundancy of visual keys, \kv\ stores a compact subset of representative keys and reconstructs the full key sequence before attention. This reduces persistent KV storage while avoiding permanent removal of keys that may become important at later decoding steps.

Formally, let $\mathcal{M}_\ell \subseteq \{1,\dots,n\}$ denote the indices of retained keys at layer $\ell$ for an image with size $n$, where compressed length $|\mathcal{M}_\ell| = \lceil \ell_r n \rceil$ and $\ell_r < 1$ is the layer-specific retention ratio. The compressed key matrix is obtained via row selection: 
\[ \tilde{\mathbf{K}}_\ell \in \mathbb{R}^{\lceil \ell_r n \rceil \times d_{\text{head}}}.\] 
To recover the full key, we employ a lightweight reconstruction network $f_{\theta_\ell}(\cdot)$ implemented as a two-layer MLP:
\[ \hat{\mathbf{K}}_\ell = f_{\theta_\ell}(\tilde{\mathbf{K}}_\ell), \quad \hat{\mathbf{K}}_\ell \in \mathbb{R}^{n \times d_{\text{head}}}. \]
Because visual keys exhibit strong sequence-level redundancy, a lightweight two-layer MLP can reconstruct the full key sequence from retained representatives. The mask $\mathcal{M}_\ell$ and reconstructor $f_{\theta_\ell}$ are jointly optimized with reconstruction, BCE mask, and retention-ratio losses, selecting keys that support accurate recovery while meeting the target compression budget. At inference, $\mathcal{M}_\ell$ and $\theta_\ell$ are fixed and reused for layer-wise key compression and reconstruction.

\paragraph{Sequence-Level Low-Rank Compression for Values.}

Values are directly aggregated into the attention output and carry token-specific visual content. As a result, pruning value rows can remove localized semantic information. Instead of selecting a subset of value tokens, \kv\ compresses values along the sequence dimension using PCA. This preserves dominant sequence-level variation while allowing all vision tokens to contribute to the compressed representation.

Formally, let $\mathbf{V}_\ell \in \mathbb{R}^{n \times d_{\text{head}}}$ denote the value matrix at layer $\ell$. We first compute a sequence-wise mean $\boldsymbol{\mu}_\ell \in \mathbb{R}^{d_\text{head}}$ and center the values as:
\[ \mathbf{V}_\ell^{\text{center}} = \mathbf{V}_\ell - \mathbf{1}_n \boldsymbol{\mu}_\ell^\top. \]
We then perform PCA along the sequence dimension and retain the top $\lceil \ell_r n \rceil$ principal components. This yields a projection matrix $\mathbf{U}_\ell \in \mathbb{R}^{\lceil \ell_r n \rceil \times n}$ and a compressed value representation:
\begin{equation*}
\tilde{\mathbf{V}}_\ell = \mathbf{U}_\ell \mathbf{V}_\ell^{\text{center}}, \quad \tilde{\mathbf{V}}_\ell \in \mathbb{R}^{\lceil \ell_r n \rceil \times d_{\text{head}}}.
\end{equation*}
During standard attention, the full-length value sequence is reconstructed by linear re-projection:
\begin{equation*}
\hat{\mathbf{V}}_\ell = \mathbf{U}_\ell^\top \tilde{\mathbf{V}}_\ell + \mathbf{1}_n \boldsymbol{\mu}_\ell^\top, \quad \hat{\mathbf{V}}_\ell \in \mathbb{R}^{n \times d_{\text{head}}}.
\end{equation*}

The projection matrix $\mathbf{U}_\ell$ is obtained via singular value decomposition of centered value activations collected from a multimodal dataset. Retaining the top $\lceil \ell_r n \rceil$ components gives the optimal rank-$\lceil \ell_r n \rceil$ approximation under the Frobenius norm. Unlike token pruning, this sequence-level projection does not discard entire value rows; instead, each token contributes to the compressed representation through the learned basis. 

Algorithm~\ref{alg:kv_pyramid_train} in Appendix~\ref{sec:kv_pyramid_train} summarizes the training procedure for key masks, key reconstructors, and value PCA bases. Appendix~\ref{app:recon_matrix} compares alternative reconstruction designs and supports our hybrid MLP-for-keys/PCA-for-values configuration.

\subsection{Inference}
\label{sec:inference}

At inference time, all learned masks, key reconstructors, PCA bases, and centering vectors are fixed. Following the layer-wise compression tolerance observed in Section~\ref{sec:layer_wise}, \kv\ uses a pyramid retention schedule: earlier layers retain more visual KV information, while deeper layers are compressed more aggressively.

\paragraph{Prefilling Stage}
During prefilling, \kv\ initializes a compressed KV cache for vision tokens. At each layer $\ell$, visual keys are compressed by the learned mask $\mathcal{M}_\ell$, and visual values are centered and projected onto the sequence-level PCA basis $\mathbf{U}_\ell$. Only the compressed visual keys and values are stored in the persistent cache. Text-token KV states are left unchanged, preserving the original language-cache behavior of the backbone VLM.

\paragraph{Decoding Stage}
During decoding, \kv\ reconstructs full-length visual KV representations before attention. Keys are recovered by the lightweight reconstructor $f_{\theta_\ell}$, while values are restored by linear re-projection using $\mathbf{U}_\ell$ and $\boldsymbol{\mu}_\ell$. The reconstructed visual KV tensors are concatenated with the uncompressed text KV tensors and passed to the original attention module. Thus, \kv\ reduces persistent visual KV storage without modifying the backbone attention interface. Since attention is still computed over a reconstructed full-length visual sequence, the model can adapt to shifting vision token importance during generation. Appendix~\ref{app:ablation_static_dynamic_compression} shows that this recoverable decoding strategy better preserves the full-cache attention pattern than a static compressed variant.

\section{Experiments}
\label{sec:experiments}

We evaluate \kv\ on accuracy preservation, robustness across image and video tasks, cross-backbone generality, and inference efficiency. In all experiments, \kv\ reduces stored visual KV length by about 60\% on average using a pyramid layer-wise schedule, while competing baselines are configured to a comparable 55\%-65\% storage reduction for fair comparison. Additional ablations on layer-wise compression ratios and input resolution are provided in Appendices~\ref{app:compression_ratio} and~\ref{app:resolution}.

\subsection{Image-based Evaluation on VLMEvalKit}
\label{sec:vlmevalkit}

Table~\ref{tab:VLMEvalKit} reports VLMEvalKit~\cite{duan2024vlmevalkit} results on MME~\cite{zhang2024mme}, MMMU~\cite{yue2023mmmu}, COCO captioning~\cite{lin2015microsoftcococommonobjects}, MMBench~\cite{liu2024mmbench}, LLaVABench~\cite{liu2023llava, liu2023improvedllava}, and HallusionBench~\cite{guan2024hallusionbench}, covering perception, reasoning, captioning, instruction following, hallucination robustness, and multi-image understanding. Across five VLM backbones, \kv\ closely matches the full-cache baseline while storing only compressed visual KV states. Compared with static pruning baselines, \kv\ is more stable across models and tasks, where irreversible token removal causes large degradation. These results show that recoverable asymmetric compression preserves visual information better than token pruning.

\begin{table*}[t!]
\fontsize{27}{36}\selectfont
\setlength{\tabcolsep}{10pt}
\centering
\caption{Performance comparison of \kv\ and other baselines on VLMEvalKit across VLMs.}
\vspace{-5pt}
\label{tab:VLMEvalKit}
\begin{threeparttable}

\scalebox{0.3}{
 \begin{tabular}{l|lcccccc} \specialrule{1pt}{0pt}{2pt} &\multirow{4}{*}{~~~VLMs {\huge *}} & \multicolumn{1}{c}{MME} & \multicolumn{1}{c}{MMMU}& \multicolumn{1}{c}{COCO}& \multicolumn{1}{c}{MMBench}& \multicolumn{1}{c}{LLaVABench}  & \multicolumn{1}{c}{HallusionBench}\\ 
 
 \specialrule{1pt}{2pt}{2pt} \multirow{5}{*}
 
\multirow{3}{*}{\makebox[30pt][c]{\rotatebox{90}{\fontsize{24}{22}\selectfont Qwen2}}} 

&\cellcolor{magenta!10}~~~Baseline: all KV  &\cellcolor{magenta!10}1687.70 &\cellcolor{magenta!10}0.42 &\cellcolor{magenta!10}14.28 &\cellcolor{magenta!10}0.82 &\cellcolor{magenta!10}80.80 &\cellcolor{magenta!10}60.12 
\\ \cline{2-8} 
&\cellcolor{magenta!10}~~~\kv\ &\cellcolor{magenta!10}1634.22 &\cellcolor{magenta!10}\textbf{0.42} &\cellcolor{magenta!10}14.40 &\cellcolor{magenta!10}\textbf{0.83} &\cellcolor{magenta!10}74.80 &\cellcolor{magenta!10}\textbf{60.99} 
\\ 
&\cellcolor{magenta!10}~~~FastV 
&\cellcolor{magenta!10}1498.46 &\cellcolor{magenta!10}0.42 &\cellcolor{magenta!10}15.00 &\cellcolor{magenta!10}0.68 &\cellcolor{magenta!10}55.50 &\cellcolor{magenta!10} 58.68
\\
&\cellcolor{magenta!10}~~~SparseVLMs 
&\cellcolor{magenta!10}1637.74 & \cellcolor{magenta!10}0.41 & \cellcolor{magenta!10}14.28 & \cellcolor{magenta!10}0.81 & \cellcolor{magenta!10}75.30 & \cellcolor{magenta!10}60.74
\\
&\cellcolor{magenta!10}~~~Prumerge 
&\cellcolor{magenta!10}1637.64 & \cellcolor{magenta!10}0.42 & \cellcolor{magenta!10}14.28 & \cellcolor{magenta!10}0.81 & \cellcolor{magenta!10}74.70 & \cellcolor{magenta!10}60.74
\\

\specialrule{1pt}{2pt}{2pt}
 
\multirow{3}{*}{\makebox[30pt][c]{\rotatebox{90}{\fontsize{24}{20}\selectfont Qwen2.5}}} 
&\cellcolor{green!10}~~~Baseline: all KV  &\cellcolor{green!10}1685.80 &\cellcolor{green!10}0.20 &\cellcolor{green!10}15.40 &\cellcolor{green!10}0.80 &\cellcolor{green!10}85.90 &\cellcolor{green!10}65.60
\\ 
\cline{2-8} 
&\cellcolor{green!10}~~~\kv
&\cellcolor{green!10}1635.80 &\cellcolor{green!10}\textbf{0.20}&\cellcolor{green!10}14.40 &\cellcolor{green!10}0.79 &\cellcolor{green!10}\textbf{83.70}  &\cellcolor{green!10}\textbf{62.20}
\\
&\cellcolor{green!10}~~~FastV
&\cellcolor{green!10}155.30 &\cellcolor{green!10}0.14&\cellcolor{green!10}5.90 &\cellcolor{green!10}0.00 &\cellcolor{green!10}12.90 &\cellcolor{green!10}25.60
\\
&\cellcolor{green!10}~~~SparseVLMs 
&\cellcolor{green!10}1637.64 & \cellcolor{green!10} 0.20 & \cellcolor{green!10}15.40 & \cellcolor{green!10}0.80 & \cellcolor{green!10}80.20 & \cellcolor{green!10}60.74
\\
&\cellcolor{green!10}~~~Prumerge 
&\cellcolor{green!10}1637.64 & \cellcolor{green!10}0.20 & \cellcolor{green!10}15.40 & \cellcolor{green!10}0.80 & \cellcolor{green!10}78.20 & \cellcolor{green!10}60.74
\\ 

\specialrule{1pt}{2pt}{2pt} 

\multirow{3}{*}{\makebox[30pt][c]{\rotatebox{90}{\fontsize{24}{100}\selectfont Qwen3}}}  
&\cellcolor{blue!10}~~~Baseline: all KV  
& \cellcolor{blue!10}1735.60 & \cellcolor{blue!10}0.52 & \cellcolor{blue!10}15.70 & \cellcolor{blue!10}0.80 & \cellcolor{blue!10}74.80 & \cellcolor{blue!10}53.90
\\ 
\cline{2-8} & 
\cellcolor{blue!10}~~~\kv
&\cellcolor{blue!10}\textbf{1708.49} & \cellcolor{blue!10}\textbf{0.48} & \cellcolor{blue!10}\textbf{15.80} & \cellcolor{blue!10}0.78 & \cellcolor{blue!10}\textbf{73.80} & \cellcolor{blue!10}53.70
\\
&\cellcolor{blue!10}~~~FastV 
&\cellcolor{blue!10}561.25 & \cellcolor{blue!10}0.35 & \cellcolor{blue!10}8.01 & \cellcolor{blue!10}0.39 & \cellcolor{blue!10}22.30 & \cellcolor{blue!10}47.83
\\
&\cellcolor{blue!10}~~~SparseVLMs 
&\cellcolor{blue!10}1706.44 & \cellcolor{blue!10}0.34 & \cellcolor{blue!10}15.60 & \cellcolor{blue!10}0.80 & \cellcolor{blue!10}72.00 & \cellcolor{blue!10}47.03
\\
&\cellcolor{blue!10}~~~Prumerge 
&\cellcolor{blue!10}1560.20 & \cellcolor{blue!10}0.48 & \cellcolor{blue!10}15.70 & \cellcolor{blue!10}0.73 & \cellcolor{blue!10}73.40 & \cellcolor{blue!10}53.80
\\ 
 
\specialrule{1pt}{2pt}{2pt} 
 
\multirow{3}{*}{\makebox[30pt][c]{\rotatebox{90}{\fontsize{24}{100}\selectfont \shortstack{LLaVA-Mistral}}}}
&\cellcolor{red!10}~~~Baseline: all KV  
& \cellcolor{red!10}1458.12 & \cellcolor{red!10}0.40
& \cellcolor{red!10}14.39 & \cellcolor{red!10}0.68 & \cellcolor{red!10}67.80 & \cellcolor{red!10}40.80
\\
\cline{2-8} 
& \cellcolor{red!10}~~~\kv
& \cellcolor{red!10}\textbf{1458.10} & \cellcolor{red!10}\textbf{0.41} & \cellcolor{red!10}14.42 & \cellcolor{red!10}0.67 & \cellcolor{red!10}\textbf{67.80} & \cellcolor{red!10}40.80
\\
& \cellcolor{red!10}~~~FastV
& \cellcolor{red!10}1440.69 & \cellcolor{red!10}0.37 & \cellcolor{red!10}15.24 & \cellcolor{red!10}0.68 & \cellcolor{red!10}67.70 & \cellcolor{red!10}40.80
\\
&\cellcolor{red!10}~~~SparseVLMs 
&\cellcolor{red!10}1339.93 & \cellcolor{red!10}0.39 & \cellcolor{red!10}15.60 & \cellcolor{red!10}0.64 & \cellcolor{red!10}67.80 & \cellcolor{red!10}45.11
\\
&\cellcolor{red!10}~~~Prumerge 
&\cellcolor{red!10}1457.62 & \cellcolor{red!10}0.39 & \cellcolor{red!10}14.30 & \cellcolor{red!10}0.68 & \cellcolor{red!10}66.50 & \cellcolor{red!10}48.15
\\ 
\specialrule{1pt}{2pt}{2pt} 
 
\multirow{3}{*}{\makebox[30pt][c]{\rotatebox{90}{\fontsize{24}{100}\selectfont \shortstack{LLaVA-Llama}}}}
&\cellcolor{cyan!10}~~~Baseline: all KV 
& \cellcolor{cyan!10}1515.40 & \cellcolor{cyan!10}0.46 & \cellcolor{cyan!10}14.70 & \cellcolor{cyan!10}0.72 & \cellcolor{cyan!10}66.30 & \cellcolor{cyan!10}37.90 
\\
\cline{2-8}
& \cellcolor{cyan!10}~~~\kv
& \cellcolor{cyan!10}\textbf{1533.40} & \cellcolor{cyan!10}\textbf{0.60} & \cellcolor{cyan!10}\textbf{14.70} & \cellcolor{cyan!10}\textbf{0.71} & \cellcolor{cyan!10}\textbf{66.00} & \cellcolor{cyan!10}49.40 
\\ 
& \cellcolor{cyan!10}~~~FastV & \cellcolor{cyan!10}1525.65 & \cellcolor{cyan!10}0.46 & \cellcolor{cyan!10}14.50 & \cellcolor{cyan!10}0.69 & \cellcolor{cyan!10}64.30 & \cellcolor{cyan!10}43.60 
\\
&\cellcolor{cyan!10}~~~SparseVLMs 
&\cellcolor{cyan!10}1504.79 & \cellcolor{cyan!10}0.45 & \cellcolor{cyan!10}14.40 & \cellcolor{cyan!10}0.69 & \cellcolor{cyan!10}65.20 & \cellcolor{cyan!10}50.00
\\
&\cellcolor{cyan!10}~~~Prumerge 
&\cellcolor{cyan!10}1526.40 & \cellcolor{cyan!10}0.49 & \cellcolor{cyan!10}14.30 & \cellcolor{cyan!10}0.71 & \cellcolor{cyan!10}65.20 & \cellcolor{cyan!10}52.79
\\
\specialrule{1pt}{2pt}{0pt} 
\end{tabular}
}
\begin{tablenotes}
    \scriptsize
    \item[] Evaluated models: \texttt{Qwen2-VL-7B-Instruct}, \texttt{Qwen2.5-VL-7B-Instruct}, \texttt{Qwen3-VL-8B-Instruct},
    \item[] \texttt{llava-v1.6-mistral-7b-hf}, \texttt{llama3-llava-next-8b-hf}. 
\end{tablenotes}
\end{threeparttable}\vspace{-10pt}
\end{table*}
\subsection{Video Understanding Evaluation}
\label{sec:video_evaluation}

To evaluate \kv\ beyond static images, Table~\ref{tab:video} reports results on Video-MME~\cite{fu2025video} and MMBench-Video~\cite{fang2024mmbench}. Across both benchmarks, \kv\ closely matches the full-cache baseline on short, medium, and long video inputs, while remaining more stable than pruning baselines under dynamic visual contexts. These results show that recoverable visual KV compression preserves the information needed for temporally extended multimodal reasoning.
\begin{table*}[t!]
\fontsize{27}{36}\selectfont
\setlength{\tabcolsep}{10pt}
\centering
\caption{Performance comparison of KVCapsule and other baselines on video datasets across VLMs.}
\label{tab:video}
\scalebox{0.3}{
\begin{tabular}{l|lccccc}
\specialrule{1pt}{0pt}{2pt}
&
\multirow{2}{*}{}
& \multicolumn{1}{c}{MME(short)}
& \multicolumn{1}{c}{MME(medium)}
& \multicolumn{1}{c}{MME(long)}
& \multicolumn{1}{c}{MMBench(coarse)}
& \multicolumn{1}{c}{MMBench(fine)}
\\

\specialrule{1pt}{2pt}{2pt}

\multirow{3}{*}{\makebox[30pt][c]{\rotatebox{90}{\fontsize{24}{22}\selectfont Qwen2}}}
& \cellcolor{magenta!10}~~~Baseline: all KV
& \cellcolor{magenta!10}0.60
& \cellcolor{magenta!10}0.47
& \cellcolor{magenta!10}0.44
& \cellcolor{magenta!10}1.23
& \cellcolor{magenta!10}1.23
\\ \cline{2-7}

& \cellcolor{magenta!10}~~~\kv\
& \cellcolor{magenta!10}\textbf{0.60}
& \cellcolor{magenta!10}0.47
& \cellcolor{magenta!10}\textbf{0.44}
& \cellcolor{magenta!10}\textbf{1.23}
& \cellcolor{magenta!10}1.23
\\ 

& \cellcolor{magenta!10}~~~FastV
& \cellcolor{magenta!10}0.58
& \cellcolor{magenta!10}0.49
& \cellcolor{magenta!10}0.42
& \cellcolor{magenta!10}1.30
& \cellcolor{magenta!10}1.24
\\
& \cellcolor{magenta!10}~~~SparseVLMs
& \cellcolor{magenta!10}0.60
& \cellcolor{magenta!10}0.47
& \cellcolor{magenta!10}0.44
& \cellcolor{magenta!10}1.24
& \cellcolor{magenta!10}1.22

\\ 
& \cellcolor{magenta!10}~~~Prumerge
& \cellcolor{magenta!10}0.60
& \cellcolor{magenta!10}0.47
& \cellcolor{magenta!10}0.43
& \cellcolor{magenta!10}1.23
& \cellcolor{magenta!10}1.22
\\ 
\specialrule{1pt}{2pt}{2pt}
\multirow{3}{*}{\makebox[30pt][c]{\rotatebox{90}{\fontsize{24}{22}\selectfont Qwen2.5}}}
& \cellcolor{green!10}~~~Baseline: all KV
& \cellcolor{green!10}0.59
& \cellcolor{green!10}0.50
& \cellcolor{green!10}0.46
& \cellcolor{green!10}1.18
& \cellcolor{green!10}1.22
\\ \cline{2-7}

& \cellcolor{green!10}~~~\kv\
& \cellcolor{green!10}0.59
& \cellcolor{green!10}\textbf{0.50}
& \cellcolor{green!10}\textbf{0.44}
& \cellcolor{green!10}\textbf{0.90}
& \cellcolor{green!10}\textbf{1.01}
\\ 

& \cellcolor{green!10}~~~FastV
& \cellcolor{green!10}0.58
& \cellcolor{green!10}0.49
& \cellcolor{green!10}0.43
& \cellcolor{green!10}0.17
& \cellcolor{green!10}0.08
\\
& \cellcolor{green!10}~~~SparseVLMs
& \cellcolor{green!10}0.60
& \cellcolor{green!10}0.48
& \cellcolor{green!10}0.44
& \cellcolor{green!10}0.69
& \cellcolor{green!10}0.73

\\ 
& \cellcolor{green!10}~~~Prumerge
& \cellcolor{green!10}0.60
& \cellcolor{green!10}0.48
& \cellcolor{green!10}0.43
& \cellcolor{green!10}0.90
& \cellcolor{green!10}0.96
\\ 
\specialrule{1pt}{2pt}{2pt}

\multirow{3}{*}{\makebox[30pt][c]{\rotatebox{90}{\fontsize{24}{22}\selectfont Qwen3}}}
& \cellcolor{blue!8}~~~Baseline: all KV
& \cellcolor{blue!8}0.66
& \cellcolor{blue!8}0.54
& \cellcolor{blue!8}0.50
& \cellcolor{blue!8}1.49
& \cellcolor{blue!8}1.56
\\ \cline{2-7}

& \cellcolor{blue!8}~~~\kv\
& \cellcolor{blue!8}\textbf{0.66}
& \cellcolor{blue!8}\textbf{0.54}
& \cellcolor{blue!8}\textbf{0.50}
& \cellcolor{blue!8}1.49
& \cellcolor{blue!8}\textbf{1.58}
\\ 

& \cellcolor{blue!8}~~~FastV
& \cellcolor{blue!8}0.33
& \cellcolor{blue!8}0.32
& \cellcolor{blue!8}0.30
& \cellcolor{blue!8}0.56
& \cellcolor{blue!8}0.59
\\
& \cellcolor{blue!8}~~~SparseVLMs
& \cellcolor{blue!8}0.58
& \cellcolor{blue!8}0.49
& \cellcolor{blue!8}0.49
& \cellcolor{blue!8}1.20
& \cellcolor{blue!8}1.24
\\ 
& \cellcolor{blue!8}~~~Prumerge
& \cellcolor{blue!8}0.60
& \cellcolor{blue!8}0.48
& \cellcolor{blue!8}0.43
& \cellcolor{blue!8}1.50
& \cellcolor{blue!8}1.53
\\ 
\specialrule{1pt}{2pt}{2pt}
\end{tabular}
}
\end{table*}


\subsection{Efficiency Analysis}
\label{sec:efficiency_analysis}

\paragraph{Inference Latency.}
\begin{algorithm}[t!]
\caption{Fused \kv\ for Layer $\ell$}
\label{alg:fused_decode_attn}
\small
\begin{algorithmic}

\STATE \textbf{Input:}
query $\mathbf{Q}$; compressed visual KV $(\tilde{\mathbf{K}}_\ell,\tilde{\mathbf{V}}_\ell)$;
key reconstructor $\theta_\ell$; value PCA $(\mathbf{U}_\ell,\boldsymbol{\mu}_\ell)$;
scaling $\sigma$; \\ compressed length $\lceil \ell_r n \rceil$; groups $G$; rank $R$

\STATE \textbf{Output:} Attention output $Y$

\STATE \textcolor{gray}{\textit{// Step 1: Compute scores in compressed sequence space}}
\STATE $S_{\mathrm{comp}} \gets (\mathbf{Q}\tilde{\mathbf{K}}_\ell^{\top}) \cdot \sigma$
\STATE $S_{\mathrm{full}} \gets \textsc{Reshape}(S_{\mathrm{comp}}, [B, H_{kv}, G, R, \lceil \ell_r n \rceil])$

\STATE \textcolor{gray}{\textit{// Step 2: Expand to spatial resolution}}
\STATE $L_{\mathrm{img}} \gets \textsc{MatMul}(S_{\mathrm{full}}, \theta_\ell)$
\STATE $P \gets \textsc{Softmax}(L_{\mathrm{img}})$

\STATE \textcolor{gray}{\textit{// Step 3: Associative Re-projection with Transposed U}}
\STATE $W_\ell \gets \textsc{MatMul}(P, \mathbf{U}_\ell^{\top})$ 

\STATE \textcolor{gray}{\textit{// Step 4: Multiply with compressed values}}
\STATE $Y \gets \textsc{MatMul}(W_\ell,\tilde{\mathbf{V}}_\ell)+\textsc{MatMul}(P,\boldsymbol{\mu}_\ell)$

\STATE Return  $Y$
\end{algorithmic}
\end{algorithm}

As shown in Algorithm~\ref{alg:fused_decode_attn}, fused \kv\ integrates reconstruction into decode-time attention, avoiding explicit materialization of full-length visual KV tensors. For values, instead of reconstructing $\hat{\mathbf{V}}_\ell$ before attention, we use associativity:
\begin{equation*}
P(\mathbf{U}_\ell^\top \tilde{\mathbf{V}}_\ell + \boldsymbol{\mu}_\ell)
=
(P\mathbf{U}_\ell^\top)\tilde{\mathbf{V}}_\ell + P\boldsymbol{\mu}_\ell .
\end{equation*}
This projects the full-resolution attention probabilities into the compressed value space, allowing the final multiplication with the value cache to operate over $\lceil \ell_r n \rceil$ compressed value states rather than $n$ full-length vision states.

For keys, fused \kv\ first computes attention scores using the compressed keys, then expands these scores before softmax instead of materializing the full key cache. Since KV tensors are repeatedly read during decoding while scores are temporary, this reduces dominant KV memory traffic while preserving attention over the reconstructed visual sequence. The gain is algorithmic, arising from compression and associative reordering rather than hardware-specific optimization.

\begin{figure}[t!]
\centering

\begin{subfigure}[t]{0.49\columnwidth}
    \centering
    \includegraphics[width=\linewidth]{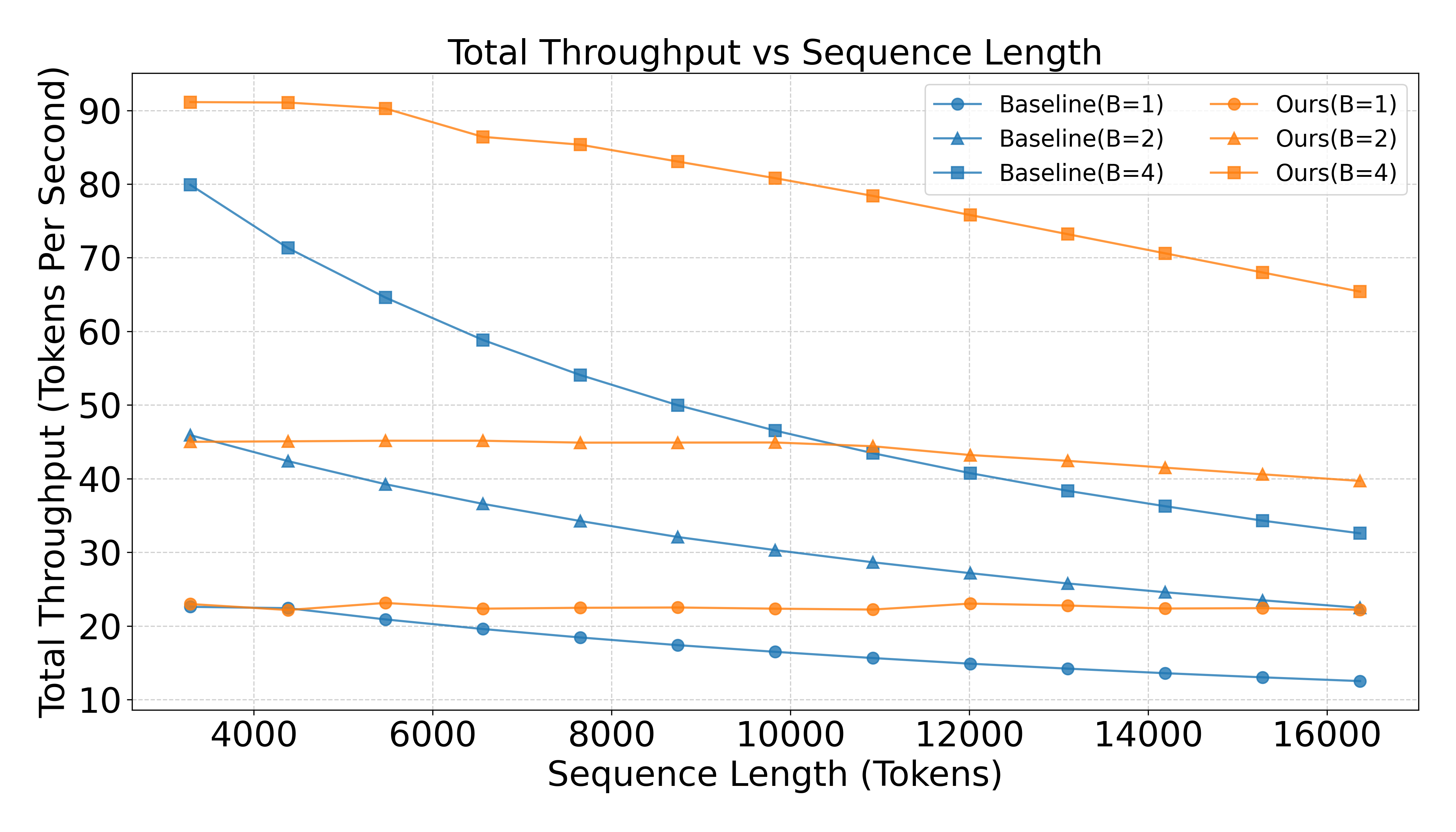}
    \caption{Throughput scaling on AMD MI250X.}
    \label{fig:tps_performance}
\end{subfigure}
\hfill
\begin{subfigure}[t]{0.48\columnwidth}
    \centering
    \includegraphics[width=\linewidth]{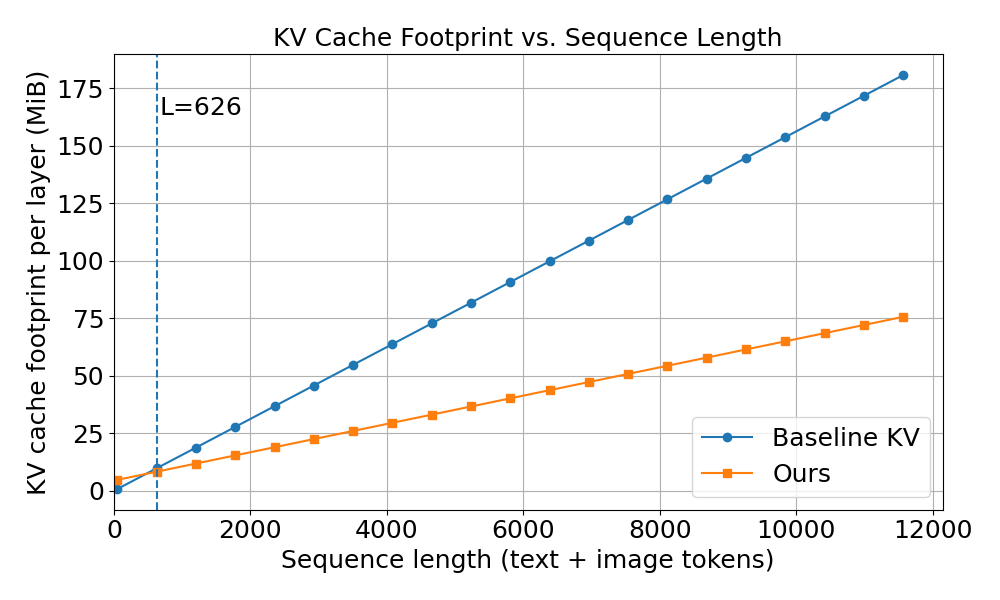}
    \caption{Persistent KV memory footprint.}
    \label{fig:theory_mem}
\end{subfigure}

\vspace{4pt}

\begin{subfigure}[t]{0.98\columnwidth}
\centering
\small
\setlength{\tabcolsep}{5.0pt}
\renewcommand{\arraystretch}{1.08}
\begin{tabular}{l|ccccccc}
\specialrule{1pt}{0pt}{2pt}
\textbf{Method / Metric (ms/token)}
& \textbf{9K}
& \textbf{10K}
& \textbf{11K}
& \textbf{12K}
& \textbf{13K}
& \textbf{14K}
& \textbf{15K} \\
\specialrule{1pt}{2pt}{2pt}

All KV 
& 74.58
& 73.23
& 76.47
& 79.69
& 82.91
& 86.05
& 89.16 \\

\kv
& \textbf{70.22}
& \textbf{67.02}
& \textbf{68.46}
& \textbf{67.69}
& \textbf{67.00}
& \textbf{67.58}
& \textbf{66.18} \\

\midrule

Latency reduction
& \textbf{5.8\%}
& \textbf{8.5\%}
& \textbf{10.5\%}
& \textbf{15.1\%}
& \textbf{19.2\%}
& \textbf{21.5\%}
& \textbf{25.8\%} \\

\specialrule{1pt}{2pt}{0pt}
\end{tabular}
\caption{Per-token latency.}
\label{tab:latency_input_length}
\end{subfigure}
\vspace{-4pt}
\caption{\textbf{Efficiency analysis of fused \kv.}
(a) Fused \kv\ improves decoding throughput over the full-cache baseline.
(b) \kv\ reduces persistent KV memory after the break-even point at $L\approx626$ tokens.
(c) For batch size 1, \kv\ reduces per-token latency across 9K-15K inputs, with relative reductions reported in the last row.}
\label{fig:efficiency_analysis}
\vspace{-8pt}
\end{figure}

Figure~\ref{fig:efficiency_analysis}(a) shows that fused \kv\ improves decoding TPS over the full-cache baseline, while Figure~\ref{fig:efficiency_analysis}(c) shows that \kv\ reduces batch-size-1 per-token latency across 9K-15K inputs, with larger reductions at longer input lengths.


\paragraph{Theoretical Memory Footprint.}
We analyze the persistent memory footprint $\mathcal{S}$ of a single Transformer layer to quantify the trade-off between KV cache savings and reconstruction overhead. For the full-cache baseline, memory scales with the stored sequence length $(c+n+t)$, including text prompt tokens, vision tokens, and generated tokens. In contrast, \kv\ stores only $\lceil \ell_r n \rceil$ compressed vision tokens, changing the dominant KV cache term from $(c+n+t)$ to $(c+\lceil \ell_r n \rceil+t)$.

\kv\ adds fixed layer-wise overhead from the key reconstructor $\theta_\ell$ and value PCA parameters $(\mathbf{U}_\ell,\boldsymbol{\mu}_\ell)$, but this overhead is independent of batch size and generated length. As shown in Figure~\ref{fig:efficiency_analysis}(b), \kv\ reaches a break-even point at $L\approx626$ tokens; beyond this point, persistent visual-KV savings outweigh reconstruction-parameter overhead. The full derivation is provided in Appendix~\ref{app:memory_analysis}.

\section{Related Work}

\textbf{Quantization-aware KV cache compression.} These methods reduce KV cache memory by lowering tensor precision. Su et al.~\cite{su2025akvq} propose an adaptive quantization strategy tailored to multimodal inputs, while Han et al.~\cite{han2025calibquant} introduce a calibration framework to retain accuracy after quantizing the KV cache. These methods operate at the element-wise level and do not reduce sequence length. In contrast, our method structurally compresses visual KV cache along the sequential axis, enabling higher memory savings without precision loss.

\paragraph{Low-rank and subspace approximations.}  
Low-rank methods compress KV tensors by projecting them into low-dimensional subspaces. Saxena et al.~\cite{saxena2024eigen} apply eigen decomposition to capture dominant token features, Zhang et al.~\cite{zhang2024lorc} propose orthogonal projections for cache reuse, and Mu et al.~\cite{mu2025sals} use subspace alignment for layer-wise KV compression. These approaches assume global low-rank structure and often require full-tensor reconstruction during decoding. Our method avoids such assumptions by leveraging localized, layer-dependent redundancy in vision sequences and reconstructing only what is needed before each attention step.

\paragraph{Attention-guided token selection and pruning.}  
Pruning methods identify and retain salient KV entries based on attention scores. Tu et al.~\cite{tu2024vl} propose vision token distillation based on average attention, Chang et al.~\cite{chang2025xkv} prune cross-modal tokens with learned gates, Pei et al.~\cite{pei2024cross} use cross-attention similarity, Jiang et al.~\cite{jiang2025purekv} apply early exit mechanisms to stop storing low-value KV pairs, Yang et al.~\cite{yang2025topv} select top-K based on relevance, and Huang et al.~\cite{huang2025aircache} explore adaptive spatial region selection. These methods generally assume that token importance remains stable throughout generation and typically prune based on prefilling or early decoding scores. Our approach challenges this assumption by empirically showing that visual attention patterns evolve during decoding and exhibit asymmetric redundancy across keys and values. We compress vision token sequences at the sequential level and reconstruct before attention computation, allowing the model to dynamically adapt its attention without being constrained by static compression decisions.

\section{Conclusion}
\label{sec:conclusion}

We presented \kv, an asymmetric visual KV cache compression framework for efficient VLM inference. \kv\ combines MLP-based key reconstruction with sequence-level PCA for values, preserving recoverable visual information while reducing persistent KV storage. Across multiple VLMs and image/video benchmarks, \kv\ closely preserves full-cache accuracy while achieving up to $2\times$ TPS improvement and $2.4\times$ persistent memory reduction. 

A current limitation is that the fixed reconstruction and PCA parameters introduce overhead, so the benefits are strongest when visual contexts are long enough for KV cache savings to dominate; our Torch-level implementation also leaves room for further gains from optimized serving kernels. 

By reducing memory demand and serving cost, \kv\ can make long-context multimodal inference more accessible and resource-efficient. As with any VLM inference system, deployment should follow the safety policies and safeguards of the underlying models. We hope this work motivates future VLM systems that natively operate on compressed multimodal KV caches for scalable inference.

\newpage
\bibliographystyle{plainnat}
\bibliography{reference}

\newpage
\appendix
\onecolumn

\section{Quantitative Analysis of Key-Value Asymmetry}
\label{app:kv_redundancy_analysis}

To strengthen our observation, we conduct a quantitative analysis on 500 randomly selected MEGABench samples. As shown in Table~\ref{tab:key_value_cosine_similarity}, keys consistently exhibit much higher inter-token cosine similarity than values across representative layers. For example, key similarity remains high from shallow to deeper layers, while value similarity stays substantially lower. This supports our observation that keys contain stronger inter-token redundancy, whereas values are more token-specific and should avoid direct token dropping. These results support our asymmetric design choice of applying structure-preserving reconstruction to keys and PCA-based subspace compression to values.
\vspace{-15pt}
\begin{table}[ht]
\centering
\caption{Inter-token cosine similarity of keys and values across representative layers, measured on 500 randomly selected MEGABench samples.}
\label{tab:key_value_cosine_similarity}
\small
\setlength{\tabcolsep}{8pt}
\renewcommand{\arraystretch}{1.12}
\begin{tabular}{lcc}
\toprule
\textbf{Layer} & \textbf{Key Similarity} & \textbf{Value Similarity} \\
\midrule
Layer 0  & 0.99 & 0.35 \\
Layer 10 & 0.87 & 0.40 \\
Layer 20 & 0.66 & 0.30 \\
\bottomrule
\end{tabular}
\end{table}
\vspace{-5pt}

\section{Hidden-Dimension Compression Is Inefficient for Visual Values}
\label{app:hidden_dim_pca}

\begin{figure}[ht]
    \centering
    \includegraphics[width=0.7\linewidth]{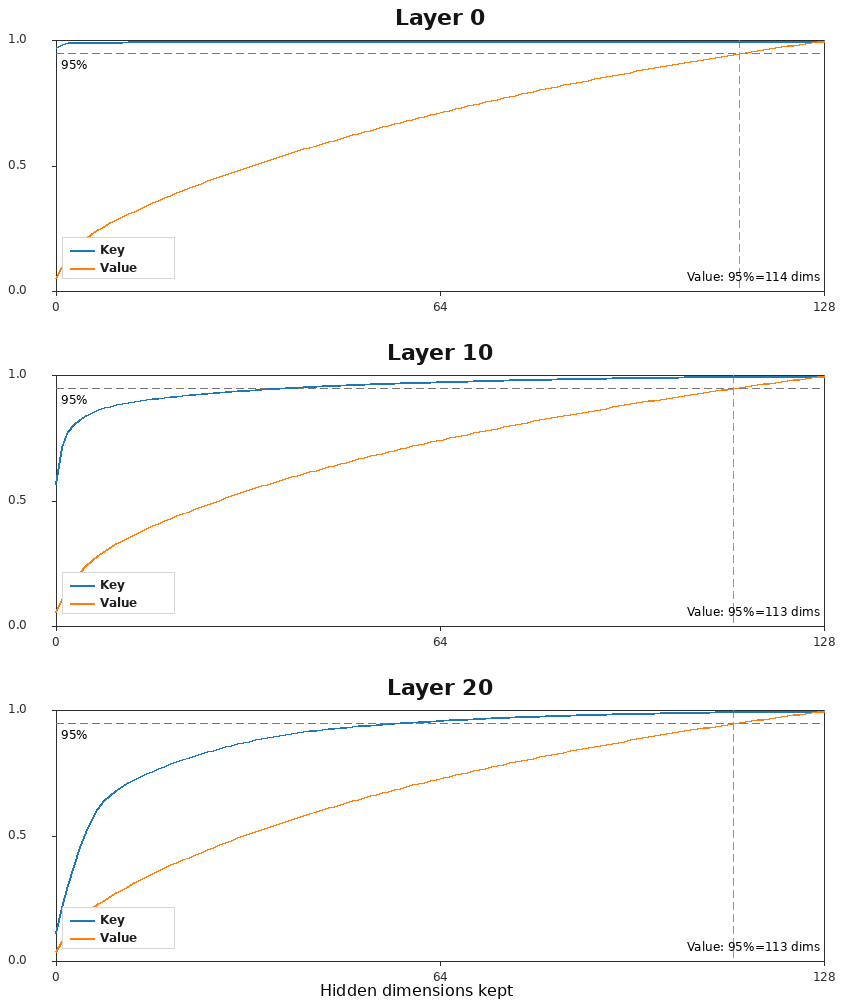}
    \caption{\textbf{Hidden-dimension PCA energy for visual KV states.}
    We compute cumulative PCA energy over feature channels for visual keys and values. Values require nearly the full hidden dimension to preserve 95\% variance across layers, suggesting that hidden-dimension compression is inefficient for visual values.}
    \label{fig:hidden_dim_pca}
\end{figure}

\begin{table}[t]
\centering
\caption{Hidden-dimension PCA ranks required to preserve variance over feature channels. Results are computed on 500 randomly sampled MEGABench examples.}
\label{tab:hidden_dim_pca_rank}
\small
\setlength{\tabcolsep}{5pt}
\renewcommand{\arraystretch}{1.12}
\begin{tabular}{l|ccc}
\toprule
\textbf{Metric} & \textbf{Layer 0} & \textbf{Layer 10} & \textbf{Layer 20} \\
\midrule
Key rank 90\%   & 1   & 15  & 38  \\
Key rank 95\%   & 1   & 37  & 57  \\
Key rank 99\%   & 3   & 89  & 97  \\
\midrule
Value rank 90\% & 102 & 99  & 100 \\
Value rank 95\% & 114 & 113 & 113 \\
Value rank 99\% & 125 & 125 & 125 \\
\midrule
Value rank 95\% fraction & 0.89 & 0.88 & 0.88 \\
Value rank 99\% fraction & 0.98 & 0.98 & 0.98 \\
Value top-10 ratio       & 0.28 & 0.32 & 0.26 \\
\bottomrule
\end{tabular}
\vspace{-6pt}
\end{table}

To examine whether visual KV states can be compressed along the hidden dimension, we compute the cumulative PCA energy of keys and values over feature channels using 500 randomly sampled MEGABench examples. As shown in Figure~\ref{fig:hidden_dim_pca} and Table~\ref{tab:hidden_dim_pca_rank}, visual values are dense along the feature dimension: preserving 95\% variance requires retaining 113-114 out of 128 dimensions across layers, and preserving 99\% variance requires 125 dimensions. This makes hidden-dimension compression inefficient for visual values.

This analysis also distinguishes \kv\ from MLA-style latent KV compression, which reduces KV memory by projecting keys and values into a compact latent feature space. In contrast, our results show that visual values are poorly compressible along the hidden dimension, while their sequence-level structure can still be exploited by PCA across vision tokens. Therefore, \kv\ preserves the hidden dimension and compresses values along the vision token sequence, reducing memory without aggressively discarding feature-channel information.

\section{Training Procedure for Layer-wise \kv\ Compression}
\label{sec:kv_pyramid_train}

The training procedure shown in Algorithm~\ref{alg:kv_pyramid_train} separates the learning of keys and values according to their different roles in attention.
For values, the PCA bases are fitted once per layer from the training split, since value compression is formulated as a fixed sequence-level low-rank projection.
For keys, the mask and reconstructor are optimized iteratively: the schedule controls the training phase and temperature $\tau$, while the mask loss encourages the selected tokens to satisfy the target retention ratio $\ell_r$.
After training, all learned components are fixed and reused during inference, so the method does not require modifying or fine-tuning the VLM backbone.

\label{app:training_algorithm}
\begin{algorithm}[ht]
\caption{Layer-wise Training Procedure for \kv}
\label{alg:kv_pyramid_train}
\small
\begin{algorithmic}
\STATE \textbf{Input:} Layer-wise KV dataset $\{\mathcal{D}_\ell\}_{\ell=1}^{L}$, retention ratio $\ell_r$
\STATE \textbf{Output:} Learned key reconstructor $\{\theta_\ell\}$ and value PCA bases $\{(\mathbf{U}_\ell,\boldsymbol{\mu}_\ell)\}$
\STATE Split $\mathcal{D}_\ell$ into training and validation sets
\STATE Initialize key reconstructor $f_{\theta_\ell}$ with retention ratio $\ell_r$
\STATE \textcolor{gray}{\textit{// Sequence-level PCA for values (once per layer)}}
\STATE $(\mathbf{U}_\ell,\boldsymbol{\mu}_\ell) \gets \textsc{FitValuePCA}(\mathcal{D}_\ell^{\mathrm{train}}, \lceil \ell_r n \rceil)$
\STATE Save $(\mathbf{U}_\ell,\boldsymbol{\mu}_\ell)$
\FOR{$e$ in Epochs}
    \STATE $(\text{phase},\tau) \gets \textsc{TrainingSchedule}(e)$
    \FOR{$(K_\ell,V_\ell) \sim \mathcal{D}_\ell^{\mathrm{train}}$}
        \STATE $\mathcal{M}_\ell \gets \textsc{GetMask}(\theta_\ell,\tau,\ell_r,\text{phase})$
        \STATE $\tilde{K}_\ell \gets K_\ell \odot \mathcal{M}_\ell$
        \STATE $\hat{K}_\ell \gets f_{\theta_\ell}(\tilde{K}_\ell)$
        \STATE $\mathcal{L} \gets \text{MSE}(\hat{K}_\ell, K_\ell) + \lambda\, \mathcal{L}_\text{mask}(\mathcal{M}_\ell, \ell_r)$
        \STATE Update $\theta_\ell$ by minimizing $\mathcal{L}$
    \ENDFOR
\ENDFOR
\end{algorithmic}
\end{algorithm}

\section{Theoretical Memory Footprint of \kv}
\label{app:memory_analysis}

We analyze the persistent memory footprint $\mathcal{S}$ of a single Transformer layer $\ell$.
In a standard Grouped-Query Attention (GQA) architecture, the baseline footprint $\mathcal{S}_{\text{base}}$ stores both keys and values for the text prompt, vision tokens, and generated tokens:
\begin{equation}
\mathcal{S}_{\text{base}}
=
2 \cdot B \cdot H_{kv} \cdot (c+n+t) \cdot d_{\text{head}} \cdot b_{kv},
\end{equation}
where $B$ is the batch size, $H_{kv}$ is the number of KV heads, $c$ is the prompt length, $n$ is the number of vision tokens, $t$ is the generation step, $d_{\text{head}}$ is the head dimension, $b_{kv}$ is the bytes per KV element.

\kv\ reduces the stored visual sequence length from $n$ to $\lceil \ell_r n \rceil$ and introduces reconstruction parameters.
The resulting memory footprint is:
\begin{equation}
\begin{aligned}
\mathcal{S}_{\text{ours}}
=
&\underbrace{
2 \cdot B \cdot H_{kv} \cdot
(c+\lceil \ell_r n \rceil+t)
\cdot d_{\text{head}} \cdot b_{kv}
}_{\text{Compressed KV Cache}} \\
&+
\underbrace{
\mathcal{S}_{\text{recon}}(\theta_\ell)
+
\mathcal{S}_{\text{PCA}}(\mathbf{U}_\ell)
}_{\text{Parameter Overhead}} .
\end{aligned}
\end{equation}

The parameter overhead consists of two components.
First, the value reconstruction uses PCA bases
$\mathbf{U}_\ell \in \mathbb{R}^{H_{kv} \times \lceil \ell_r n \rceil \times n}$,
together with the mean term $\boldsymbol{\mu}_\ell$.
Second, the key reconstructor $f_{\theta_\ell}$ is dominated by the score reconstruction matrix
$\theta_\ell \in \mathbb{R}^{H_{kv} \times \lceil \ell_r n \rceil \times n}$,
which maps compressed attention scores back to the full vision token length for the final attention calculation.
Thus, the additional overhead is mainly determined by the vision token resolution and compressed length.

This formulation shows the trade-off between reducing the dominant KV cache term and adding reconstruction parameters.
The baseline KV cache scales with $(c+n+t)$, while \kv\ replaces the vision-token term $n$ with $\lceil \ell_r n \rceil$ in the stored KV cache.
Although the reconstruction matrices introduce additional storage, this overhead is separate from the per-batch KV cache term.
Therefore, in the long-context regime where KV cache storage dominates, the compressed KV cache savings can outweigh the reconstruction overhead, leading to a lower overall memory footprint.
As illustrated in Figure~\ref{fig:theory_mem}, this results in a reduced memory growth rate and enables longer supported sequence lengths under a fixed HBM budget.

\section{Ablation on Key and Value Reconstruction Designs}
\label{app:recon_matrix}
\begin{figure}[ht]
\begin{center}
\centerline{\includegraphics[width= \columnwidth]{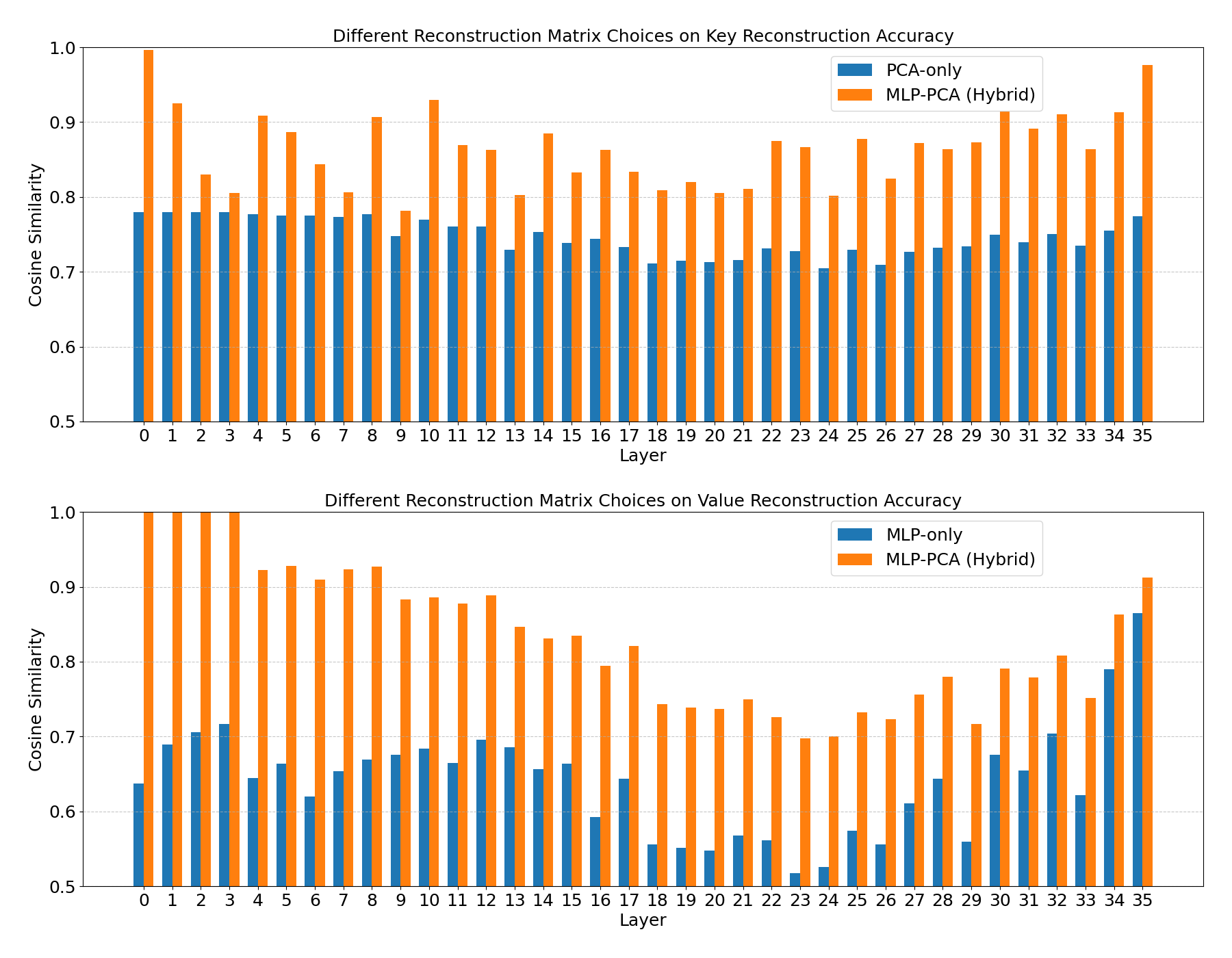}}
\caption{\textbf{Comparison of Key and Value Reconstruction Designs.} This figure illustrates the effectiveness of a hybrid MLP--PCA architecture compared to baseline methods across 36 layers of a neural network. Performance is measured by the cosine similarity between the original and reconstructed vectors (higher is better).}
\label{fig:ablation_re}
\end{center}
\end{figure}

To validate the reconstruction design in \kv, we conducted an ablation study using the \texttt{Qwen2-VL-8B-Instruct} model evaluated on 500+ randomly selected samples from the MEGABench dataset \cite{chen2025mega-bench}. The empirical evidence presented in the reconstruction plots strongly supports the hybrid design choice of utilizing a 2-layer MLP for key reconstruction and PCA for value reconstruction. By isolating these components across all 36 layers of the model, we demonstrate that this specific configuration addresses the distinct mathematical requirements of keys and values in the attention mechanism. 

By integrating these two approaches, our method achieves an optimal balance: while keys benefit from the non-linear discrimination required for selective retention of specific features, values rely on efficient preservation to maintain the integrity of their semantic content. This hybrid strategy ensures high reconstruction quality throughout the model's depth by aligning each reconstruction technique with the distinct functional roles of the KV cache components.

\section{Layer-wise Compression Ratio Sensitivity}
\label{app:compression_ratio}
\begin{figure}[ht]
\begin{center}
\centerline{\includegraphics[width= 0.7\columnwidth]{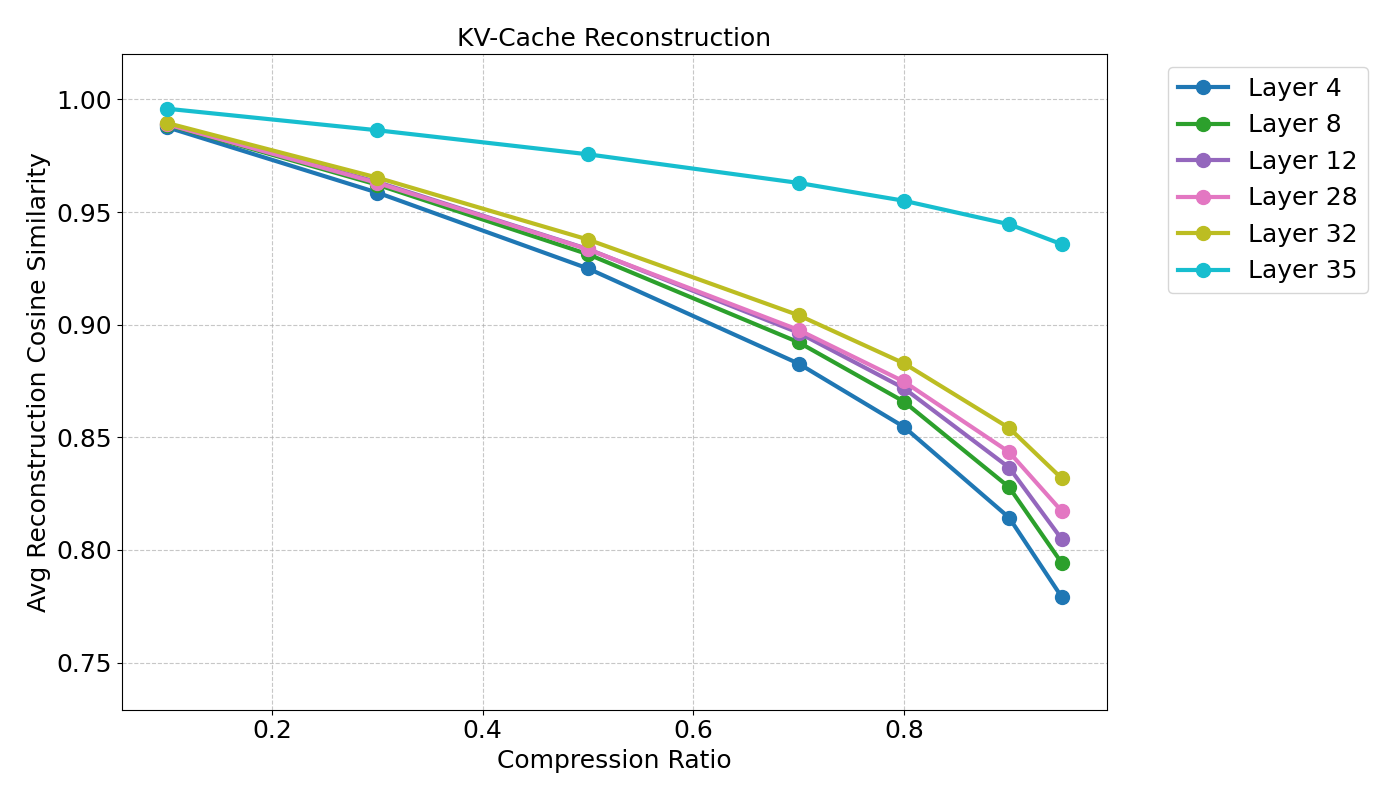}}
\caption{\textbf{KV Cache Reconstruction Accuracy vs. Compression Ratio.} The plot illustrates the relationship between compression ratio (mask ratio) and reconstruction cosine similarity across various model layers.}
\label{fig:compression_sensitivity}
\end{center}
\end{figure}

To determine the optimal distribution of memory resources across the network, we conducted a layer-wise sensitivity analysis using the \texttt{Qwen2-VL-8B-Instruct} model evaluated on 500 randomly selected samples from the MEGABench dataset. The empirical results support a pyramid compression schedule, where earlier layers require lower compression, while deeper layers tolerate more aggressive compression.

We implemented a sensitivity sweep across a representative subset of layers to measure reconstruction fidelity under varying compression ratios ranging from 0.1 to 0.95. By calculating the mean cosine similarity between the original and reconstructed tensors across different mask ratios, we identified the specific information-retention requirements of each network stage.

Sensitivity analysis in Figure~\ref{fig:compression_sensitivity} reveals that final layers are remarkably resilient to aggressive compression. For instance, Layer 35 maintains a reconstruction cosine similarity above 0.93 even at a 0.95 compression ratio, significantly outperforming middle-layer benchmarks under the same constraints.

Combined with observations in Figure~\ref{fig:kv_cos_sim}, the sequence-wise similarity heatmaps for initial layers show complex, high-contrast patterns and dense feature mappings. This suggests that foundational features lack redundancy and act as high-fidelity extractors, requiring higher retention to avoid losing critical context. As the model depth increases, the heatmaps transition from complex textures into prominent vertical and horizontal stripes. These patterns appear because certain "anchor" tokens begin to share a high degree of similarity with almost every other token in the sequence. This convergence indicates that the model has synthesized raw inputs into more stable semantic concepts.

\section{Ablation on K-side and V-side Reconstruction}
\label{app:key_value_reconstruction}
\begin{table}[ht]
\centering
\caption{Ablation study on the individual effects of key and value reconstruction in KVCapsule.}
\label{tab:kv_ablation_attention}
\small
\setlength{\tabcolsep}{8pt}
\renewcommand{\arraystretch}{1.12}
\begin{tabular}{lcc}
\toprule
\textbf{Method} & \textbf{Cosine Similarity} & \textbf{MSE} \\
\midrule
\kv\ (K only) & 0.58 & $1.07 \times 10^{-6}$ \\
\kv\ (V only) & 0.76 & $7.27 \times 10^{-7}$ \\
\kv\ & 0.60 & $1.05 \times 10^{-6}$ \\
\bottomrule
\end{tabular}
\end{table}
The K-only / V-only ablation isolates how each component affects the final attention pattern.
The results in Table~\ref{tab:kv_ablation_attention} show that attention fidelity is more sensitive to K-side compression, since approximating keys directly changes the attention-score geometry and can shift where the model attends.
In contrast, V-side compression appears less disruptive under attention-pattern metrics, but this does not mean values can be compressed aggressively: values carry the content aggregated by attention, so even small value perturbations can still degrade generation quality.
This observation motivates KVCapsule's asymmetric design: we reconstruct keys to better preserve attention geometry, while applying a more conservative PCA-based compression to values instead of hard pruning.
Overall, the ablation supports our claim that keys and values play distinct roles and should be compressed with different strategies rather than a single uniform compression rule.

\section{Ablation on Static Compression vs. Reconstruction-before-Attention}
\label{app:ablation_static_dynamic_compression}

\begin{table}[H]
\centering
\caption{Ablation study on static vs. dynamic lossy compression. We compare the final attention weights against the full-cache baseline.}
\label{tab:ablation_static_dynamic_compression}
\small
\setlength{\tabcolsep}{8pt}
\renewcommand{\arraystretch}{1.12}
\begin{tabular}{lcc}
\toprule
\textbf{Comparison} & \textbf{Cosine Similarity} & \textbf{MSE} \\
\midrule
All KV vs. \kv\ w/ reconstruction  & 0.60 & $7.74 \times 10^{-7}$ \\
All KV vs. static \kv\ w/o reconstruction  & 0.24 & $1.07 \times 10^{-6}$ \\
\bottomrule
\end{tabular}
\end{table}

We conduct an ablation study to compare dynamic lossy compression with per-step reconstruction against a static lossy compression variant where reconstruction is disabled and the compressed KV is used directly throughout decoding.
The results in Table~\ref{tab:ablation_static_dynamic_compression} show that dynamic reconstruction preserves the final attention pattern much better than the static compressed variant under the same compressed storage setting.
This supports our motivation that vision token importance and attention geometry evolve during decoding, making a fixed compressed representation brittle.
By reconstructing KV before attention, \kv\ better adapts the compressed representation to the current decoding step and reduces the mismatch from the full-cache baseline.

\section{Image Resolution Sensitivity}
\label{app:resolution}
Table~\ref{tab:resolution} reports the ablation study of \kv\ under different input image resolutions on Qwen3-VL across VLMEvalKit benchmarks. In the main paper, we provide the results with $1 \times$ resolution. Here, we also report results using the same images at $4 \times$ higher spatial resolution (four times the number of input pixels). Overall, these results indicate that \kv\ maintains a robust accuracy-efficiency trade-off under moderate resolution reduction.
\begin{table}[H]
\fontsize{27}{36}\selectfont
\setlength{\tabcolsep}{10pt}
\centering
\caption{Performance comparison of \kv\ with different resolutions on VLMEvalKit.}
\label{tab:resolution}
\begin{threeparttable}

\scalebox{0.3}{
 \begin{tabular}{l|lcccccc} \specialrule{1pt}{0pt}{2pt} &\multirow{4}{*}{~~~VLMs {\huge *}} & \multicolumn{1}{c}{MME} & \multicolumn{1}{c}{MMMU}& \multicolumn{1}{c}{COCO}& \multicolumn{1}{c}{MMBench}& \multicolumn{1}{c}{LLaVABench}  & \multicolumn{1}{c}{HallusionBench}\\ 
 
 \specialrule{1pt}{2pt}{2pt} \multirow{5}{*}
 
\multirow{2}{*}{\makebox[30pt][c]{\rotatebox{90}{\fontsize{22}{20}\selectfont Qwen3}}} 

&\cellcolor{blue!10}~~~Baseline: all KV  
& \cellcolor{blue!10}1735.60 & \cellcolor{blue!10}0.52 & \cellcolor{blue!10}15.70 & \cellcolor{blue!10}0.80 & \cellcolor{blue!10}74.80 & \cellcolor{blue!10}53.90
\\ \cline{2-8} 
&\cellcolor{blue!10}~~~\kv: 1x resolution
&\cellcolor{blue!10}1708.49 & \cellcolor{blue!10}0.48 & \cellcolor{blue!10}15.80 & \cellcolor{blue!10}0.78 & \cellcolor{blue!10}73.80 & \cellcolor{blue!10}53.70
\\ 
&\cellcolor{blue!10}~~~\kv: 4x resolution
&\cellcolor{blue!10}1740.78 & \cellcolor{blue!10}0.54 & \cellcolor{blue!10}13.12 & \cellcolor{blue!10}0.84  & \cellcolor{blue!10}76.80 & \cellcolor{blue!10}63.40
\\ 
&\cellcolor{blue!10}~~~FastV 
&\cellcolor{blue!10}561.25 & \cellcolor{blue!10}0.35 & \cellcolor{blue!10}8.01 & \cellcolor{blue!10}0.39 & \cellcolor{blue!10}22.30 & \cellcolor{blue!10}47.83
\\ 

\specialrule{1pt}{2pt}{0pt} 
\end{tabular}
}
\end{threeparttable}\vspace{-10pt}
\end{table}


\end{document}